\documentclass[11pt,a4paper]{article}
\usepackage[final]{emnlp2021}

\usepackage{tabu} 
\usepackage{bbm}
\usepackage{transparent}

\usepackage{caption}
\usepackage{multirow}
\usepackage{CJKutf8}
\usepackage{nccmath}

\usepackage{balance}
\usepackage{pifont}

\usepackage{times}
\usepackage{latexsym}
\usepackage{tikz}

\usepackage{booktabs}
\usepackage{diagbox,tabularx}
\usepackage{graphicx}
\usepackage{subfig}
\usepackage{xspace}

\usepackage{fontawesome}
\usepackage{bm}
\usepackage{dsfont}
\usepackage{amsmath}
\usepackage{amssymb}
\usepackage{amsthm}
\usepackage{algorithm}

\newcommand{\notes}[1]{}

\theoremstyle{definition}

\theoremstyle{plain}

\newcommand{\ith}[1]{\ensuremath{i^{{th}}}}

\newcommand{\goesto}{\ensuremath{\rightarrow}\xspace}

\newcount\permx
\newcount\permy
\def\permdot#1#2{
\permx=#1 \advance\permx by-1
\permy=#2 \advance\permy by-1
\psframe[fillcolor=black, fillstyle=solid]
(\permx,\permy)(#1, #2)
}

\newcommand{\argmax}{\operatornamewithlimits{\mathbf{argmax}}}

\newcommand{\boxnum}[1]{{\setlength{\fboxsep}{1pt}\raisebox{1pt}{\hspace{1pt}\fbox{\tiny #1}\hspace{1pt}}}}
\newcommand{\ind}[1]{\ensuremath{_{\kern-0.5pt\boxnum{#1}}}}

\newcommand{\vecx}{\ensuremath{{\mathbf{x}}}\xspace}
\newcommand{\vecy}{\ensuremath{{\mathbf{y}}}\xspace}

\newcommand{\smallnt}[1]{\ensuremath{_{\mbox{\tiny PP}}}\xspace}

\newcommand{\pseudocode}{Algorithm}
\floatname{algorithm}{\pseudocode}

\iffalse

\else

\fi

\newcommand{\bmtheta}{\ensuremath{\bm{\theta}}}
\newcommand{\thetafull}{\ensuremath{\bm{\theta}^\text{full}}}
\newcommand{\thetawk}{\ensuremath{\bm{\theta}^{\text{wait-$k$}}}}
\newcommand{\pfull}{\ensuremath{p_{\text{full}}}}
\newcommand{\pwk}{\ensuremath{p_{\text{wait-$k$}}}}
\newcommand{\waitk}{\ensuremath{\text{wait-$k$}}\xspace}
\newcommand{\HR}{\ensuremath{\mathit{HR}}\xspace}

\definecolor{tgreen}{rgb}{0,0.7,0.3}
\definecolor{torange}{rgb}{1,0.65,0}
\mathchardef\mhyphen="2D 

\usepackage{microtype}

\newcommand{\tabincell}[2]{\begin{tabular}{@{}#1@{}}#2\end{tabular}}

\title{Improving Simultaneous Translation by Incorporating Pseudo-References with Fewer Reorderings}

\author{
  Junkun Chen $^{1}$\thanks{$^*$Equal contribution. $^\dagger$Currently at Columbia University.} \quad
  Renjie Zheng $^{2 *}$ \quad
  Atsuhito Kita $^{1 \dagger}$ \quad
  Mingbo Ma $^{2}$ \quad
  Liang Huang $^{1,2}$
\\
  $^{1}$Oregon State University, Corvallis, OR, USA \\
  $^{2}$Baidu Research, Sunnyvale, CA, USA \\
  \texttt{chenjun2@oregonstate.edu, renjiezheng@baidu.com} \\
}

\date{}

\begin{document}

\begin{CJK}{UTF8}{gbsn}
\maketitle
\begin{abstract}

Simultaneous translation is vastly different from 
full-sentence translation,
in the sense that it starts translation before the source sentence ends,
with only a few words delay.
However, due to the lack of large-scale, 
high-quality
simultaneous translation datasets,
most such 
systems are still trained on conventional 
full-sentence bitexts. 
This 
is far from ideal for the simultaneous scenario due to the
abundance of
unnecessary long-distance reorderings in those bitexts.
We propose a novel method that rewrites the target side of existing full-sentence corpora
into simultaneous-style translation.
Experiments on Zh$\rightarrow$En and Ja$\rightarrow$En simultaneous translation show substantial improvements
(up to +2.7 BLEU)
with the addition of these generated pseudo-references.

\end{abstract}

\section{Introduction}

Simultaneous translation, which starts translation before the source sentence ends, 
is substantially more challenging than full-sentence translation due to partial observation of the (incrementally revealed) source sentence. 
Recently, it has witnessed great progress
thanks to fixed-latency policies (such as \waitk)
\cite{ma+:2019} and adaptive policies \cite{gu+:2017,arivazhagan+:2019}. 

However, all 
state-of-the-art simultaneous translation models are trained
on conventional parallel text 
which involve many unnecessary long-distance reorderings
 \cite{birch+:2009,braune+:2012}; 
 see Fig.~\ref{fig:idea} for an example.
The simultaneous translation models trained using 
these parallel sentences
will learn to either make bold hallucinations
(for fixed-latency policies) or
introduce long delays (for adaptive ones).
Alternatively, 
one may want to use transcribed corpora from
professional simultaneous interpretation
 \cite{matsubara+:2002,bendazzoli+:2005,neubig2018naist}.
These data are more monotonic in word-order,
but they are all very small in size due to the high cost of
data collection
(e.g., the NAIST one \cite{neubig2018naist} has
only $387k$ target words).
More importantly, simultaneous  interpreters 
tend to summarize
and inevitably make many mistakes \cite{shimizu2014collection, xiong2019dutongchuan,zheng2020fluent}
due to the high cognitive load
and intense time pressure during interpretation
\cite{camayd2011cognitive}.

\begin{figure}[t]
\hspace{-0.2cm}\resizebox{0.49\textwidth}{!}{
\setlength{\tabcolsep}{1.5pt}
\renewcommand{\arraystretch}{1.}
\begin{tabu}{c | l l l l l l l l l l l}

\toprule

\rowfont{\small}
& \textit{zh\=onggu\'o} & \textit{de} & \textbf{\textcolor{orange}{\textit{xīb\`u}}} & \textbf{\textcolor{red}{\textit{y\v{o}u}}} & \textbf{\textcolor{blue}{\textit{h\v{u}ndu\=o}}} & \textbf{\textcolor{purple}{\textit{g\=ao}}} & \textbf{\textcolor{purple}{\textit{sh\=an}}} & &\\[-0.1cm]
\tabincell{c}{Source\\ Input} & 中国 & 的 & \textbf{\textcolor{orange}{西部}} & \textbf{\textcolor{red}{有}} & \textbf{\textcolor{blue}{很多}} & \textbf{\textcolor{purple}{高}} & \textbf{\textcolor{purple}{山}}   \\[-0.2cm]
 \rowfont{\small\it}
 & china & 's & \textbf{\textcolor{orange}{west}} & \textbf{\textcolor{red}{have}} & \textbf{\textcolor{blue}{many}} & \textbf{\textcolor{purple}{big}} & \textbf{\textcolor{purple}{mountain}}\!\!\!\!\! & \\
\midrule \midrule
Gold-Ref & \multicolumn{10}{l}{\textbf{\textcolor{red}{there \ are}} \  \textbf{\textcolor{blue}{many}} \ \textbf{\textcolor{purple}{big}} \ \textbf{\textcolor{purple}{mountains}} \ in \ \textbf{\textcolor{orange}{western}} \ china }\\
\midrule \midrule
\multirow{3}{*}{Pseudo-Refs}& \multicolumn{1}{c}{{\small \it (wait-1)}} &  china & 's & \textbf{\textcolor{orange}{west}} & \textbf{\textcolor{red}{has}} & \textbf{\textcolor{blue}{many}} & \textbf{\textcolor{purple}{big}} & \textbf{\textcolor{purple}{mtns}}\\
& \multicolumn{2}{c}{{\small\it (...wait-2...)}} & the & chinese & \textbf{\textcolor{orange}{west}} & \textbf{\textcolor{red}{has}} & \textbf{\textcolor{blue}{many}} & \textbf{\textcolor{purple}{big}} & \textbf{\textcolor{purple}{mtns}}    \\ 
& \multicolumn{3}{c}{{\small\it (...wait-3...)}} & \textbf{\textcolor{orange}{western}} & china & \textbf{\textcolor{red}{has}} & \textbf{\textcolor{blue}{many}} & \textbf{\textcolor{purple}{big}} & \textbf{\textcolor{purple}{mtns}} &   \\ 
& \multicolumn{4}{c}{{\small\it (...wait-4...)}} & there & are & \textbf{\textcolor{blue}{many}} & \textbf{\textcolor{purple}{big}} & ...\\
\bottomrule
 \end{tabu}
}
\caption{
Example of unnecessary reorderings in the bitext which can force the model to anticipate aggressively, along with
the ideal pseudo-references with different \waitk policies. 
Larger $k$ improves fluency but sacrifices latency 
(pseudo-refs with $k\!\geq\! 4$ are identical to the original reference). (mtns: mountains)
\label{fig:idea}
}
\end{figure}

How can we combine the merits of both types of data,
and obtain a large-scale, more monotonic parallel corpora for simultaneous
translation?
We propose a  simple and effective technique to generate pseudo-references
with fewer reorderings;
see the ``Pseudo-Refs'' in Fig.~\ref{fig:idea}.
While previous work \cite{he+:2015} addresses 
this problem via language-specific 
hand-written rules, 
our technique can be easily adopted to any  language pairs without
using extra data or expert linguistic knowledge.
Training with these generated pseudo references can reduce anticipations during training 
and result in fewer hallucinations in decoding and lower latency.
We make the following contributions:
\begin{itemize}
\setlength{\itemsep}{-0.05cm}
	\item We propose a method to generate
	 pseudo-references 
	which are \textit{non-anticipatory} and \textit{semantic preserving}.
	\item We propose two metrics to quantify the
	 anticipation rate in the pseudo-references
	 and the hallucination rate in the hypotheses.
	\item Our pseudo-references lead to  substantial improvements
	(up to $+2.7$ BLEU) on Zh$\rightarrow$En and Ja$\rightarrow$En simultaneous translation.
\end{itemize}

\section{Preliminaries}

We briefly review  full-sentence neural translation and the \waitk policy in simultaneous translation.

\begin{figure}[!tb]
  \centering
  \includegraphics[width=7.5cm]{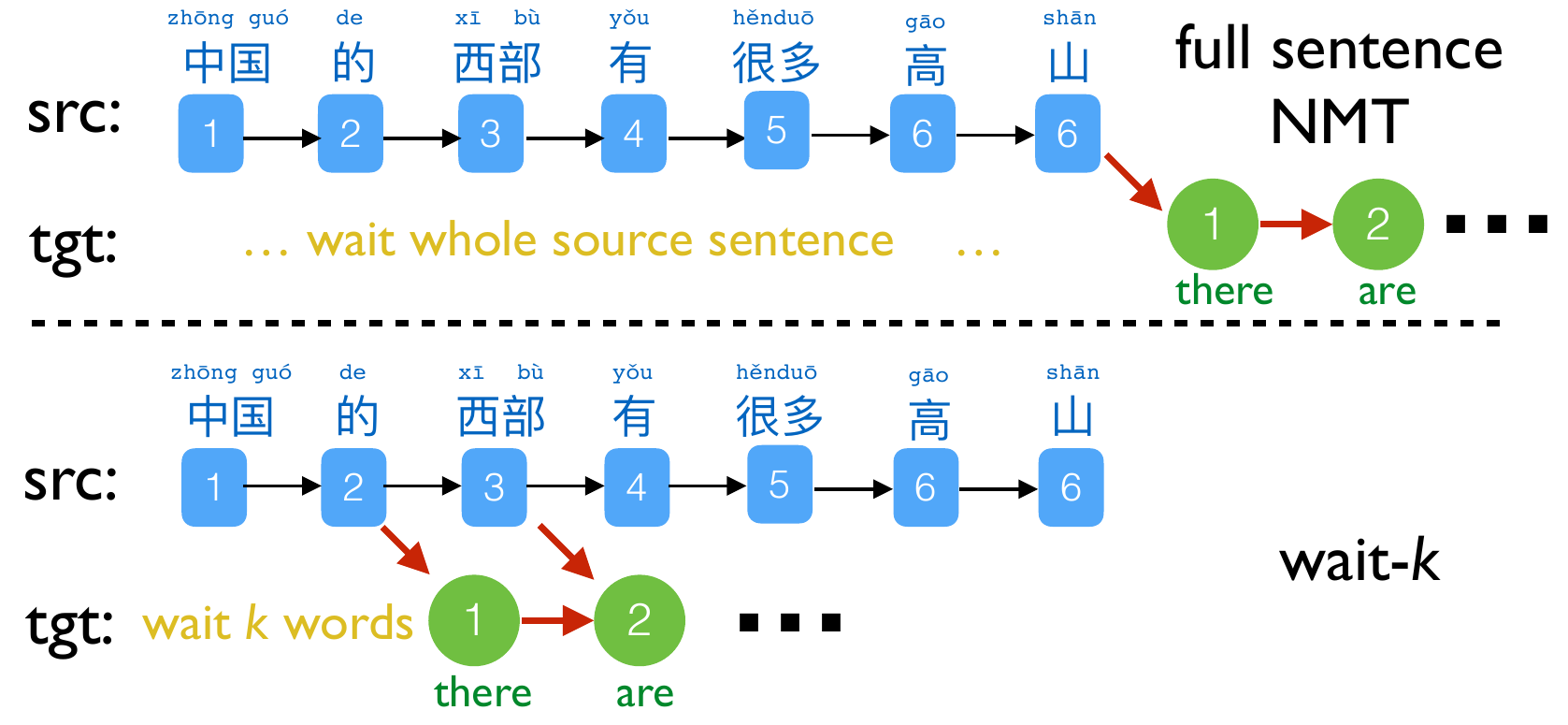}
  \captionof{figure}{Full-sentence 
  vs.~simultaneous (\waitk) MT.}
  \label{fig:pref2pref}
\end{figure}

\smallskip
\noindent\textbf{Full-Sentence NMT}
uses a Seq2seq framework
(Fig.~\ref{fig:pref2pref})
where the  encoder processes the source sentence $\vecx=(x_1,x_2,...,x_m)$  into a sequence of hidden states.
A decoder sequentially generates a target sentence $\vecy=(y_1,y_2,...,y_n)$ conditioned on those hidden states and previous predictions:
\begin{align*}
  \hat{\vecy} = \argmax_{\vecy}\pfull(\vecy \mid \vecx;\thetafull)\\
  \pfull(\vecy \mid \vecx;\bmtheta) = \prod\nolimits_{t=1}^{|\vecy|}p(y_t \mid \vecx, \vecy_{<t};\bmtheta) 
\end{align*}
The model is trained as follows:
\begin{align}
  \thetafull = \argmax_{\bmtheta} \prod\limits_{(\vecx,\vecy^*) \in D}\pfull(\vecy^* \mid \vecx;\bmtheta)
  \label{eq:seq2seq_train}
\end{align}

\noindent\textbf{Simultaneous Translation} 
 translates concurrently with the (growing)
source sentence,
so \citet{ma+:2019} propose the \waitk policy
 (Fig.~\ref{fig:pref2pref}) following a simple, 
fixed schedule 
that commits one target word on receiving each new 
source word, after an initial wait of $k$ source words.
Formally, the prediction of $\vecy$ for
a trained \waitk model is
\begin{equation}
\begin{aligned}
  &\pwk(\vecy\!\mid\!\vecx;\bmtheta)\!=\! 
\!\!\!\!  &\prod\nolimits_{t=1}^{|\vecy|}\! p(y_t \!\mid\!\vecx_{<t+k}, \vecy_{<t};\bmtheta)
\label{eq:pref2pref_train}
\end{aligned}
\end{equation}
where the \waitk model is trained as follows
  \begin{equation*}
  \thetawk=\argmax_{\bmtheta}\!\! \prod\limits_{(\vecx,\vecy^*) \in D}\pwk(\vecy^* \mid \vecx;\bmtheta).
  \end{equation*}
This way, the model learns to implicitly anticipate at testing time, though not always correct 
(e.g.,   in Fig.~\ref{fig:pref2pref}, after seeing $x_1 x_2$=``中国 的'' (China 's), output $y_1$=``there'') .
The decoder generates the target sentence 
$\hat{\vecy}$  
with $k$ words behind source sentence $\vecx$:
  \begin{equation*}
\hat{y}_t\! =\! \argmax_{y_t}\pwk(y_t \mid \vecx_{<t+k}, \hat{\vecy}_{<t};\thetawk)
\label{eq:pref2pref_decode}
\end{equation*}

\label{sec:prelim}



\section{Pseudo-Reference Generation}

\label{sec:Method}

Since the \waitk models
are  trained on conventional full-sentence bitexts,
their performance is hurt by unnecessary
long-distance reorderings between the source and target sentences.
For example, the training sentence pair in Fig.~\ref{fig:pref2pref}, 
a wait-2 model learns to 
output $y_1$=\textit{``there''}  after observing 
$x_1 x_2$=``中国 的'' ({\it china 's})
which 
seems to induce a
good anticipation (``{中国 的}...'' $\leftrightarrow$ ``{\em There ...}''), but it
 could be 
a wrong hallucination in many other contexts (e.g., ``中国 的 \underline{街道 很 挤}'' $\leftrightarrow$ ``{\em Chinese streets are crowded}'', not ``{\em There ...}'').
Even for adaptive policies \cite{gu+:2017,arivazhagan+:2019, zheng+zheng+:2019}, 
the model only learns 
a higher latency policy (wait till $x_4$=``有'') by training on the  example in Fig.~\ref{fig:pref2pref}.
As a result, training-time \waitk models
tend to do wild hallucinations \cite{ma+:2019}.

To solve this problem, we propose to generate
pseudo-references which are \textit{non-anticipatory}
under a specific simultaneous translation policy
by the method introduced in Section \ref{sec:generate}.
Meanwhile, we also propose to use BLEU score to filter the generated pseudo-references
to guarantee that they are
\textit{semantic preserving} in Section \ref{sec:filter}.

\subsection{Generating Pseudo-References with Test-time Wait-$k$}
\label{sec:generate}

To generate \textit{non-anticipatory} pseudo-references
under a \waitk policy,
we propose to use the full-sentence NMT model  $\thetafull$  (Eq.~\ref{eq:seq2seq_train})
which is {\em not} trained to anticipate,
but decode with a \waitk policy.
This combination is called {\em test-time \waitk} \cite{ma+:2019},
which is unlikely to hallucinate since the full
source content is always available during training.
Although here the full-sentence
model  $\thetafull$ only has access to the partially
available source words $\bm{x}_{<t+k}$,
it can still enforce fluency because $\hat{y}_t$  relies on the decoded  target-side prefix $\hat{\bm{y}}_{<t}$ (Eq.~\ref{eq:pref2pref_train}).
Formally, the generation of pseudo-references is:
\begin{align*}
\tilde{\vecy}^{*} = \argmax_{\vecy}\pwk(\vecy \mid \vecx;\thetafull)
\end{align*}

Fig.~\ref{fig:idea} shows the pseudo-references with different \waitk policies
($k=1..4$). Note that $k=1$ or $2$ results in non-idiomatic translations, and larger $k$ leads to more fluent pseudo-references, which converge to the original reference with $k\geq 4$.
The reason is that in each \waitk policy, each target word $\hat{y}_t$ only rely on 
observed source words ($\vecx_{<t+k}$).

To further improve the quality of the pseudo-references generated by test-time \waitk,
we propose to select better pseudo-references
by using beam search. 
Beam search usually improves translation quality 
but its application to simultaneous translation 
is non-trivial, where output words are committed on the fly \cite{zheng2019speculative}.
However, for pseudo-reference generation, unlike simultaneous translation decoding,
we can simply adopt conventional off-line beam search algorithm since
the source sentence is completely known.
A larger beam size will generally give better results,
but make anticipations more likely to be retained
if they are correct and reasonable.
To trade-off the expectations of quality and monotonicity,
we choose beam size $b=5$ in this work.

\subsection{Translation Quality of Pseudo-References}
\label{sec:filter}

\begin{figure}[!tb]
  \centering
  
    \resizebox{\linewidth}{!}{
      \includegraphics{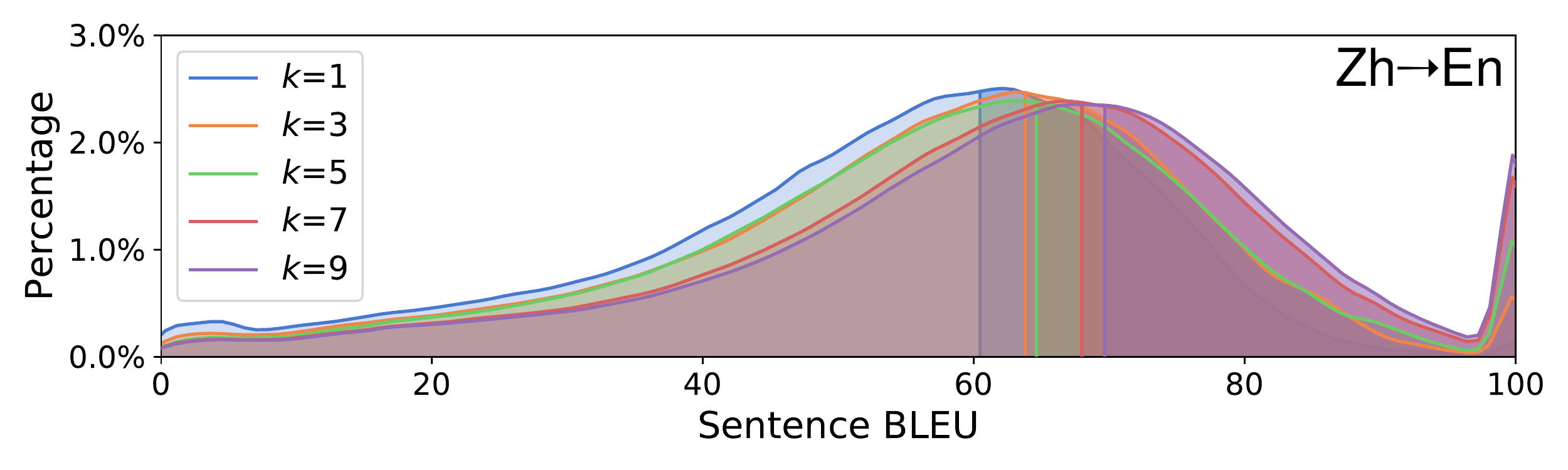}
    }
    \resizebox{\linewidth}{!}{
      \includegraphics{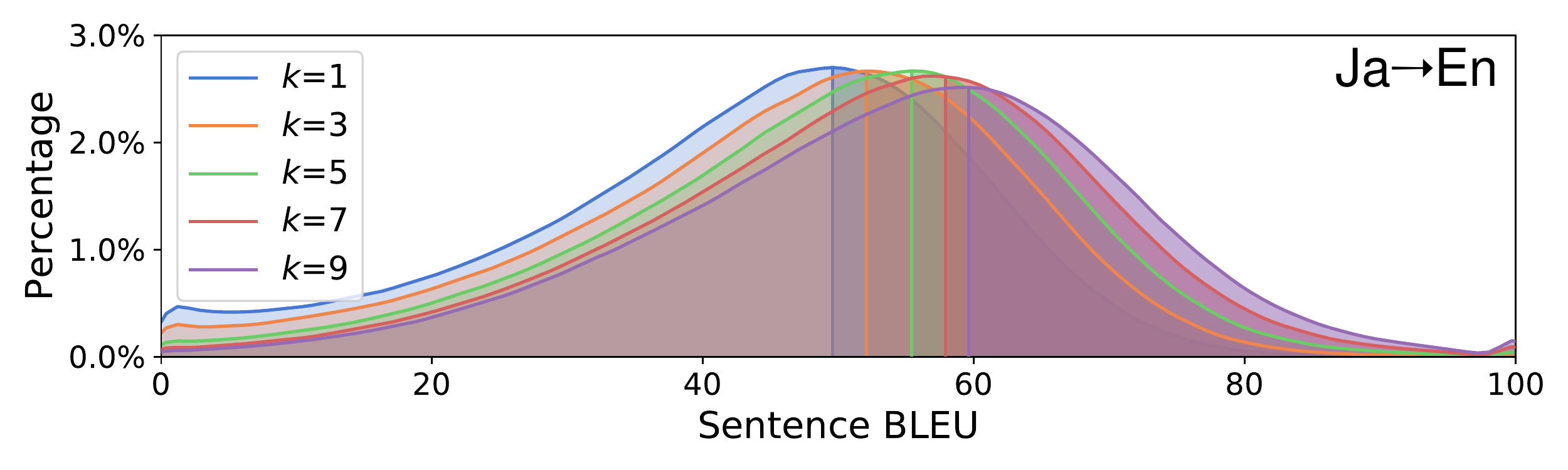}
    }
  \caption{Sentence-level BLEU distributions of Pseudo-Refs using \waitk policies for Zh$\rightarrow$En and Ja$\rightarrow$En, respectively.
  The parts to the right of the vertical lines indicate the top 40\% references in terms of BLEU in each distribution.}
  \label{fig:bleu_dist}
\end{figure}

We can use sentence-level BLEU score to filter out 
low quality pseudo-references.
Fig.~\ref{fig:bleu_dist}
shows the sentence level
BLEU distributions
of the pseudo-references generated with different 
\waitk policies. 
As $k$ increases, the translation qualities are better
since more source prefixes can be observed during decoding.
The obvious peak at the BLEU=$100$ on Zh$\rightarrow$En
denotes those pseudo-references which are identical to
the original ones. 
Those original references are probably 
already non-hallucinatory or correspond to very short
source sentences (e.g.~shorter than~$k$).
The figure shows that even for wait-1 policy, 
around 40\% pseudo-references can achieve BLEU score
above 60.

\section{Anticipation \& Hallucination Metrics}

\subsection{Anticipation Rate of (Pseudo-)References}

During the training of a simultaneous translation model,
an anticipation happens when
a target word is generated before
the corresponding source word is encoded.
To identify the anticipations, we need the word alignment between the parallel sentences.

\begin{figure}[t]
  \centering
  
    \resizebox{0.8\linewidth}{!}{
      \includegraphics{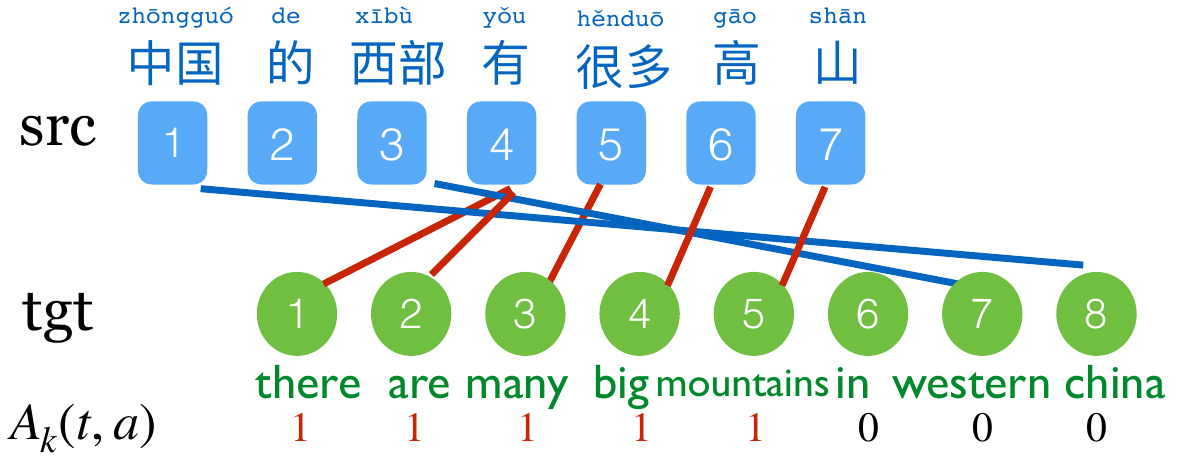}
    }
  \caption{An example  word alignment and the wait-1 policy.
  The red and blue lines indicate  the $1$-anticipated and non-anticipated alignments, resp.
  Here $AR_1$ = 5/8.
  }
  \label{fig:anticipation_exp}
\end{figure}

A word alignment $a$ between a source sentence $\vecx$
and a target sentence $\vecy$ is a set of source-target
word index pairs $(s, t)$
where
the $s^\text{th}$ source word ${x}_s$
aligns with the $t^\text{th}$ target word ${y}_t$.
In the example in Fig.~\ref{fig:anticipation_exp},
the word alignment is:
$a = \{ (1, 8), (3, 7), (4, 1), (4, 2), (5, 3), (6,4), (7,5) \}$.

Based on the word alignment $a$, we propose a new metric
called ``$k$-anticipation'' 
to detect the anticipations under \waitk policy.
Formally, a target word ${y}_t$ is $k$-anticipated
($A_k (t,a)=1$)
if it aligns to at least one source word $\vecx_s$ where $s \ge t + k$:
\begin{align*}
A_k(t,a)\!=\! \mathds{1} [ \{ (s, t) \in a\! \mid\! s \ge t + k \} \neq \varnothing]
\end{align*}

We further define the $k$-anticipation rate ($AR_k$) 
 of an $(\vecx, \vecy, a)$ triple under \waitk policy to be:
\begin{equation*}
\begin{aligned}
AR_k(\vecx, \vecy,a)\!=\!
\frac{1}
{|\vecy|}
\sum\nolimits_{t=1}^{|\vecy|}
A_k(t,a)
\end{aligned}
\end{equation*}

\subsection{Hallucination Rate of Hypotheses}

The goal of reducing the anticipation rate during
the training of a simultaneous translation model  is to avoid
hallucination at testing time.
Similar to the anticipation metric introduced in 
the previous section,
we define another metric to quantify
the number of hallucinations
in decoding.
A target word ${\hat{y}}_t$ is a {\em hallucination} if it
can not be aligned to any source word. 
Formally, based on word alignment $a$,
whether target word ${\hat{y}}_t$ is 
a hallucination is
\begin{align*}
H(t, a)\!=\! \mathds{1} [ \{ (s, t) \in a \} = \varnothing ]
\end{align*}
We further define hallucination rate \HR as 
\begin{equation*}
\begin{aligned}
\HR(\vecx, \mathbf{\hat{y}}, a)\!=\!
\frac{1}
{|\mathbf{\hat{y}}|}
\sum\nolimits_{t=1}^{|\mathbf{\hat{y}}|}
H(t, a)
\end{aligned}
\end{equation*}
To avoid non-faithful contextual alignments, we use IBM Model 1 \cite{brown+:1993} for \HR.

\section{Experiments}


\paragraph{Dataset and Model}

We conduct the experiments on two language pairs 
Zh$\rightarrow$En and Ja$\rightarrow$En.
We use NIST corpus (2M pairs) for Zh$\rightarrow$En as training set,
and NIST 2006 and NIST 2008 as dev and test set,
which contains 616 and 691 sentences with 4 English references respectively.
We also collected a set of references annotated by
human interpreters with sight-interpreting\footnote{Sight interpreting refers to (real-time) oral translation of written text.
It is considered as a special variant of simultaneous interpretation
but with better translation quality.}
for the test set.
For Ja$\rightarrow$En translation,
we use ASPEC corpus (3M pairs).
Following  \citet{morishita+:2019},
we only use the first 1.5M parallel sentences 
and discard the rest noisy data.
We use the dev and test datasets in ASPEC with 1,790 and 1,812 pairs.
We preprocess the data with Mecab \cite{kudo+:2004} as the word segmentation tool and Unidic \cite{den+:2007} as its dictionary.
Consecutive Japanese tokens which only contain Hiragana characters are combined to reduce the redundancy.

The full-sentence model is trained on the original training set.
We use \textit{fast\_align} \cite{dyer+etal+:2013} as 
the word aligner (Model 2 for anticipation and Model 1 for hallucination) and
train it on the training set. 
All the datasets are tokenized with BPE \cite{sennrich+:2016}. 
We implement \waitk policies 
on base Transformer \cite{vaswani+:2017} 
following \citet{ma+:2019} for all experiments.

\paragraph{Results}

We compare the performance of \waitk models 
trained on three different settings:
(i) original training references only;
(ii) original training references with all Pseudo-Refs;
(iii) original training references with top 40\% Pseudo-Refs in sentence-level BLEU.

\subparagraph{Chinese-to-English}

\begin{table}[th]\centering 
  \setlength{\tabcolsep}{0.3em}
  \renewcommand{\arraystretch}{0.6}
  \resizebox{\linewidth}{!}{%
  \begin{tabular}{lrrrrrr|r}
  \toprule
  \multicolumn{2}{l}{(4-reference BLEU)} &  $k$=1  & $k$=3  & $k$=5 & $k$=7 & $k$=9  &Avg.$\triangle$ \\ 
  \midrule
  Training- & BLEU $\uparrow$ &29.7 &32.1 &34.2 &35.6 &37.6  \\
  Refs (*)  & $\HR \% \downarrow$ & \textit{8.4} & \textit{7.8} & \textit{6.4} & \textit{6.0} & \textit{5.8}  \\
  \midrule
  *+100\%  & BLEU $\uparrow$ & 31.8 & 32.6 & 35.9 & 37.9 & \textbf{39.4} & \textcolor{tgreen}{$+$1.7 (\ \ 5.0\%)} \\
  Pseudo-Refs  & $\HR \% \downarrow$ & \textit{\textbf{5.5}} & \textit{7.4} & \textit{5.4} & \textit{5.2} & \textit{\textbf{4.6}} &\textcolor{torange}{$-$1.3 (18.9\%)} \\
  \midrule
  *+Top 40\%  & BLEU $\uparrow$ & \textbf{32.3} & \textbf{34.3} & \textbf{36.4} & \textbf{38.4} & 38.8 &\textcolor{tgreen}{ $+$2.2 (\ \ 6.5\%)}\\
  Pseudo-Refs  & $\HR \% \downarrow$  & \textit{5.9} & \textit{\textbf{5.8}} & \textit{\textbf{5.3}} & \textit{\textbf{5.1}} & \textit{5.3} &\textcolor{torange}{$-$1.4 (20.3\%)} \\
  \bottomrule 
  \end{tabular}
  }
  \caption{BLEU scores and hallucination rates (\HR) of Zh$\rightarrow$En  \waitk models on the test set against the original 4 references. (Full-sentence BLEU: 39.9).
  }
  \label{tab:bleu}
\end{table}

\begin{table}[th]\centering 
  \setlength{\tabcolsep}{0.2em}
  \renewcommand{\arraystretch}{0.6}
  \resizebox{\linewidth}{!}{%
  \begin{tabular}{lrrrrr|r}
  \toprule
  (single-reference BLEU)  & $k$=1  & $k$=3  & $k$=5 & $k$=7 & $k$=9  &Avg.$\triangle$ \\ 
  \midrule
  Training-Refs (*)   &10.9 &12.1 &13.0 &13.7 &13.8  \\
  \midrule
  *+Top 40\% Pseudo-Refs  & \textbf{12.6} & \textbf{14.2} & \textbf{13.9} & \textbf{14.2} & \textbf{14.1} &\textcolor{tgreen}{ $+$1.1 (7.5\%)}\\
  \bottomrule 
  \end{tabular}
  }
  \caption{BLEU scores of Zh$\rightarrow$En  \waitk models on the test set, taking human sight interpretation as reference. 
  }
  \label{tab:bleu-w-monotonic}
  \vspace{-.2cm}
\end{table}

\begin{figure}[!tb]
  \centering
   \hspace{-0.4cm} \resizebox{!}{2.9cm}{\includegraphics{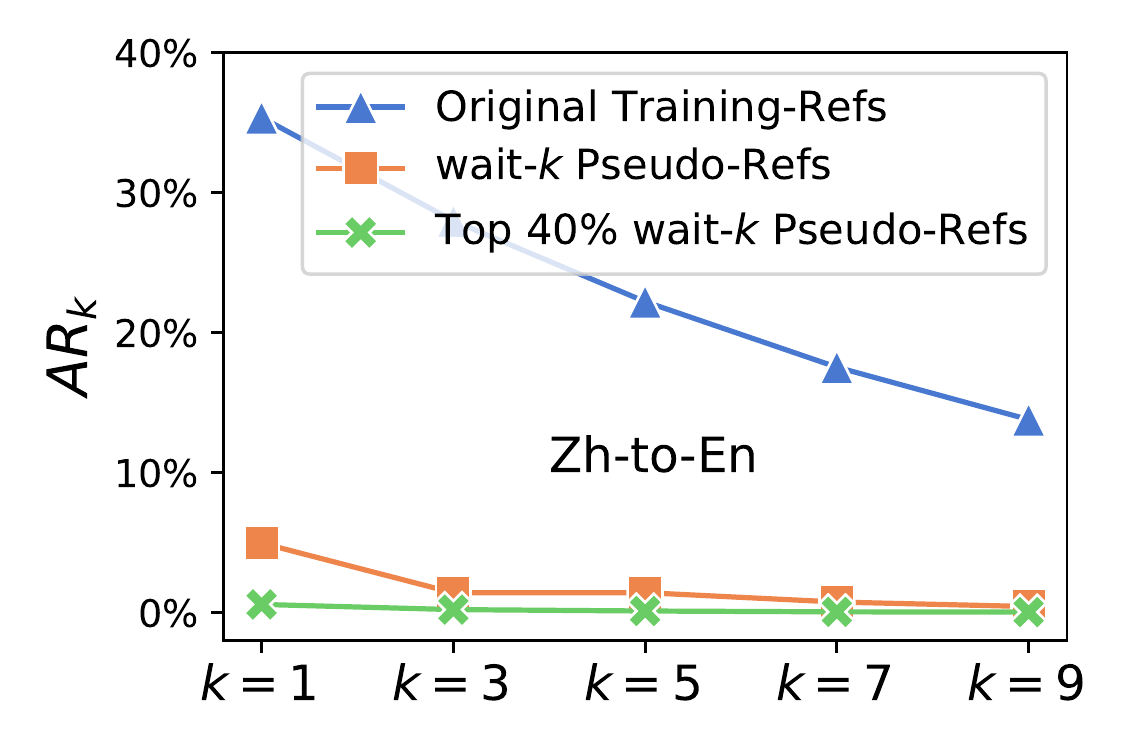}} 
    \hspace{-.35cm}
    \resizebox{!}{2.85cm}{\includegraphics{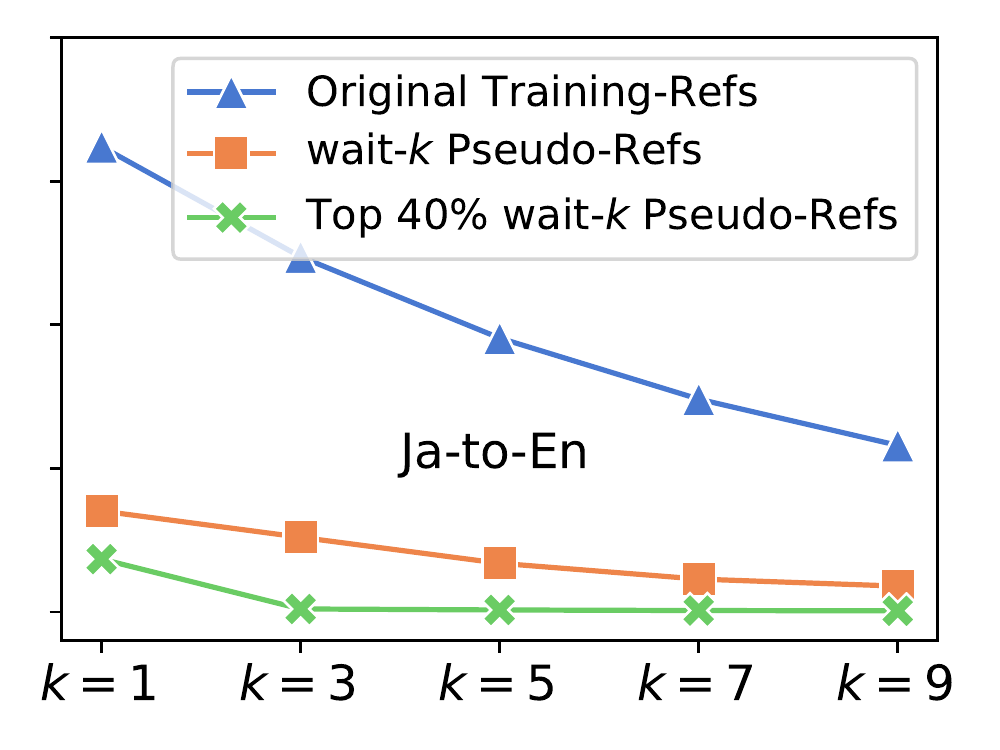}} 
    
    \label{fig:zh-en}
  \caption{$k$-Anticipation rates ($\mathit{AR}_k$) of gold training references and Pseudo-Refs with various $k$. 
       The top 40\% Pseudo-Refs are selected in terms of BLEU.}
  \label{fig:k-anticipation}
\end{figure}

Table \ref{tab:bleu} shows the results
of Zh$\rightarrow$En translation.
Compared with using original references only,
adding Pseudo-Refs substantially improves the
translation quality and reduces hallucination rate.
The filtered $40\%$ Pseudo-Refs achieve the best results 
except $k = 9$. 
Fig.~\ref{fig:k-anticipation} shows that
the generated Pseudo-Refs can significant 
reduce the $k$-anticipation rate compared with 
the original training references, especially for smaller $k$.
As shown in Table \ref{tab:bleu-w-monotonic},
if taking the human sight-interpreting result as a single reference,
the  improvement is more salient than evaluated on the standard 4 references
(+7.5\% vs.~+6.5\%),
which confirms that our method tend to translate in a ``{\em syntactic linearity}'' fashion like
human sight and simultaneous interpreters \cite{ma2019effect}.

\begin{figure*}[!h]\centering  \footnotesize

    \resizebox{!}{1.cm}{
    \hspace{-10pt}
    \setlength{\tabcolsep}{1.5pt}
    \renewcommand{\arraystretch}{0.5}
    \begin{tabu}{c | l l l l l l l l l l l l l l l l l l l}
    \toprule
    \rowfont{\small}
    & & \textbf{\textcolor{blue}{\textit{zh\=onggu\'o}}}  & \textbf{\textcolor{blue}{\textit{r\`ush\`i}}} & \textbf{\textcolor{blue}{\textit{yǐh\`ou}}} &, & \textit{zh\=ong} & \textit{m\v ei} & \textit{li\v ang} & \textit{gu\' o} & \textit{ji\=ang}\\
    \tabincell{c}{Training\\Source Input} & ...  &\textbf{\textcolor{blue}{中国}} &\textbf{\textcolor{blue}{入世}}  &\textbf{\textcolor{blue}{以后}} &, & 中 & 美 & 两 & 国 &将 & ... &\multicolumn{4}{c}{\textbf{(a) Training Example}}\\
    & &\textbf{\textcolor{blue}{china}}  & \textbf{\textcolor{blue}{entry wto}} &\textbf{\textcolor{blue}{after}} &, & china  & USA  & two & country & will &  & \\

    \rowfont{\small}
    \midrule
(a)    Gold Training-Ref &... & \textbf{\textcolor{red}{the}} & \textbf{\textcolor{red}{two}} & \textbf{\textcolor{red}{countires}} & will & ... & \textbf{\textcolor{blue}{after}} & \textbf{\textcolor{blue}{china}} & \textbf{\textcolor{blue}{'s}} & \textbf{\textcolor{blue}{entry}} & \textbf{\textcolor{blue}{into}} & \textbf{\textcolor{blue}{the}} & \textbf{\textcolor{blue}{wto}}  &. \\
    \midrule
(a')    wait-$3$ Pseudo-Ref &... & \textbf{\textcolor{blue}{after}} &\textbf{\textcolor{blue}{china}} &\textbf{\textcolor{blue}{'s}} &\textbf{\textcolor{blue}{accession}} & \textbf{\textcolor{blue}{to}} & \textbf{\textcolor{blue}{the}} & \textbf{\textcolor{blue}{wto}} & , & \textbf{\textcolor{tgreen}{china}} &\textbf{\textcolor{tgreen}{and}} & \textbf{\textcolor{tgreen}{the}}  & \textbf{\textcolor{tgreen}{united}} &\textbf{\textcolor{tgreen}{states}} &will &...\\ 

    \bottomrule
    \end{tabu}
    }
    \label{fig:case_study_train_instance}\\

    \resizebox{!}{1.0cm}{
    \centering
    \hspace{-10pt}
    \setlength{\tabcolsep}{1.5pt}
    \renewcommand{\arraystretch}{0.5}
    \begin{tabu}{c | l l l l l l l l l l l l l l l l l l l}
    \toprule
    \rowfont{\small}
    &  & & \textbf{\textcolor{blue}{\textit{f\=engzh\=ong}}}  & \textbf{\textcolor{blue}{\textit{h\`ou}}} &   & \textit{sh\v oush\`u} & \textit{q\v ud\'e} & \textit{yu\'anm\v an} & \textit{ch\'engg\=ong} &。\\
    
    \tabincell{c}{Dev\\Source Input} &\textbf{\textcolor{blue}{29@@}} &\textbf{\textcolor{blue}{5}}  &\textbf{\textcolor{blue}{分钟}}  &\textbf{\textcolor{blue}{后}}  & , & 手术 & 取得 & 圆满 & 成功 &。&\multicolumn{6}{c}{\textbf{(b) Dev-set Decoding Results}} \\
    
    &  &  &\textbf{\textcolor{blue}{minutes}} & \textbf{\textcolor{blue}{after}}  &  & surgery & achieve & complete & success & .\\
    \rowfont{\small}
    \midrule
(b)    Only Training-Refs & & & & \textbf{\textcolor{red}{the}} & \textbf{\textcolor{red}{two}} & \textbf{\textcolor{red}{countries}} & had & a & complete & success & in & the & operation & \textbf{\textcolor{blue}{after}} &\textbf{\textcolor{blue}{2@@}} &\textbf{\textcolor{blue}{95}}  &\textbf{\textcolor{blue}{minutes}}&. \\
    \midrule
(b')    + top 40\% Pseudo-Refs& & & &\textbf{\textcolor{blue}{2@@}} &\textbf{\textcolor{blue}{95}} &\textbf{\textcolor{blue}{minutes}} & \textbf{\textcolor{blue}{later}} & , & the & operation & was & a & complete & success &.\\ 
    \bottomrule
    \end{tabu}
    }
   
  \caption{
  In the training example in (a),
  the gold reference anticipates ``the two countries'',
  which encourages the \waitk model trained on it to make  irrelevant hallucination
  after any temporal phrase; see the decoding example in (b).
  Training with the pseudo-reference in (a')
  fixes this problem, resulting in the correct translation in (b').
  }
  \label{fig:case_study}
  \end{figure*}
Fig.~\ref{fig:case_study} shows an example of how
the \waitk model is improved by generated Pseudo-Refs.
In this example,
the original training references actively delay the translation of adverbial clause (time). 
It makes the model learn to anticipate the subject before its appearance.
It is common in the original set.
Fig.~\ref{fig:delay} shows two other examples of generated pseudo references on Ja$\rightarrow$En and Zh$\rightarrow$En, respectively.
The generated pseudo-references are obviously more ideal than the original references.
We also show several examples of solving other avoidable anticipations in Figs.~\ref{fig:advance}--\ref{fig:diff} in the Appendix.

\begin{figure*}[!h]\centering
  \begin{CJK*}{UTF8}{min}
  \resizebox{\linewidth}{!}{
  \setlength{\tabcolsep}{1.5pt}
  \renewcommand{\arraystretch}{1}
  \begin{tabu}{c | l l l l l l l l l l l l l l l l l l l l l l l l l }
  \toprule
  
  Training &  \textbf{\textcolor{blue}{現在}} & までに   & 症例 &・ & 对照 & の  & ２０ & ペアが  & 有効 & 回答 & として &報告  & された  &。\\
  Source Input & \textbf{\textcolor{blue}{Present}}    & by &case &and & contrast & & 20 & pairs   & effective & answers & as & are &reported &.\\

  \midrule
  Gold Training-Ref & & & & \textbf{\textcolor{red}{20}} & \textbf{\textcolor{red}{pairs}}  & of & case & and & before & contrast & \textbf{\textcolor{red}{were}} & \textbf{\textcolor{red}{reported}} & as &a  &usefulness  answers  \textbf{\textcolor{blue}{by}}  \textbf{\textcolor{blue}{the}}  \textbf{\textcolor{blue}{present}} .\\
  \midrule
  wait-3 Pseudo-Ref & & & &\textbf{\textcolor{blue}{to}} &\textbf{\textcolor{blue}{the}} &\textbf{\textcolor{blue}{present}}  &, &\textbf{\textcolor{tgreen}{20}} & \textbf{\textcolor{tgreen}{pairs}} & of & cases & and & controls & \textbf{\textcolor{tgreen}{have}} & \textbf{\textcolor{tgreen}{been}} \textbf{\textcolor{tgreen}{reported}}  as  effective  answers .\\ 

  \bottomrule
  \end{tabu}
  }\\
  \end{CJK*}

  \resizebox{!}{1.0cm}{
  \setlength{\tabcolsep}{1.5pt}
  \renewcommand{\arraystretch}{1}
  \begin{tabu}{c | l l l l l l l l l l l l l l l l l l l}
  \toprule
  \rowfont{\small}
  & \textbf{\textcolor{blue}{\textit{ji\v ang zu\`o}}}  & \textbf{\textcolor{blue}{\textit{k\=aishǐ}}} & \textbf{\textcolor{blue}{\textit{qi\'an}}} &, & lǐ & p\'eng & f\=abi\v ao & ji\v anghu\`a &。\\[-0.2cm]
  \tabincell{c}{Training\\Source Input} & \textbf{\textcolor{blue}{讲座}} & \textbf{\textcolor{blue}{开始}}  & \textbf{\textcolor{blue}{前}} &, & 李 & 鹏 & 发表 & 讲话 &。\\[-0.2cm]
  &\textbf{\textcolor{blue}{lecture}}  & \textbf{\textcolor{blue}{begin}} &\textbf{\textcolor{blue}{before}} &, & li  & peng  & deliver & speech & . &  & \\
  \midrule
  Gold Training-Ref & & & & \textbf{\textcolor{red}{li}} & \textbf{\textcolor{red}{peng}} & \textbf{\textcolor{red}{made}} & \textbf{\textcolor{red}{a}} & \textbf{\textcolor{red}{speech}} &\textbf{\textcolor{blue}{before}} \textbf{\textcolor{blue}{the}}  \textbf{\textcolor{blue}{start}}  \textbf{\textcolor{blue}{of}}  \textbf{\textcolor{blue}{the}} \textbf{\textcolor{blue}{lecture}}  \textbf{\textcolor{blue}{minutes}} . \\
  \midrule
  wait-3 Pseudo-Ref & & & &\textbf{\textcolor{blue}{before}} &\textbf{\textcolor{blue}{the}} &\textbf{\textcolor{blue}{lecture}} & \textbf{\textcolor{blue}{began}} & , & \textbf{\textcolor{tgreen}{li}} \textbf{\textcolor{tgreen}{peng}}  \textbf{\textcolor{tgreen}{gave}}  \textbf{\textcolor{tgreen}{a}}  \textbf{\textcolor{tgreen}{speech}}  .\\ 

  \bottomrule
  \end{tabu}
  }
  \caption{Two examples dealing with adverbial clause delay.
  The adverbial clauses are at the end of the training references.
  This introduces anticipation during training and
  hallucination during decoding.
  }
  \label{fig:delay}
\end{figure*}

\begin{table}[t]\centering 
  \setlength{\tabcolsep}{0.3em}
  \renewcommand{\arraystretch}{0.8}
  \resizebox{\linewidth}{!}{%
  \begin{tabular}{lrrrrr|r}
  \toprule
  \multicolumn{2}{l}{(single-reference BLEU)}  & $k$=3  & $k$=5 & $k$=7 & $k$=9 &Avg.$\triangle$ \\ 
  \midrule
  Training-  & BLEU $\uparrow$ & 16.6 & 19.0  & 20.8 & 21.7 \\
  Refs (*)  &  $\HR \% \downarrow$  & \textit{10.8} & \textit{7.3} & \textit{6.5} & \textit{6.2}  \\
  \midrule
  *+100\%  & BLEU $\uparrow$  & 17.7 & 18.9 & 20.8 & 22.2 &\textcolor{tgreen}{ $+$0.3 (1.5\%)}\\
  Pseudo-Refs & $\HR \% \downarrow$  & \textit{\textbf{6.5}} & \textit{\textbf{6.2}} & \textit{\textbf{5.6}} & \textit{5.3} &\textcolor{torange}{ $-$1.4 (18.2\%)} \\
  \midrule
  *+Top 40\% & BLEU $\uparrow$ & \textbf{17.9} & \textbf{19.2} & \textbf{21.5} & \textbf{22.5} &\textcolor{tgreen}{ $+$0.6 (3.1\%)}\\
  Pseudo-Refs  & $\HR \% \downarrow$ & \textit{8.3} & \textit{7.6} & \textit{6.0} & \textit{\textbf{5.2}}  &\textcolor{torange}{ $-$0.7 (9.1\%)} \\
  \bottomrule 
  \end{tabular}
  }
  \caption{BLEU scores and \HR of Ja$\rightarrow$En \waitk models on the test set. 
  (Full-sentence: 28.4).
  }
  \label{tab:bleu-jp}
\end{table}

\vspace{-.2cm}

\subparagraph{Japanese-to-English}
Table \ref{tab:bleu-jp} shows the results of
Ja$\rightarrow$En translation task.
Japanese-to-English  simultaneous translation 
is a more difficult task due to long distance reorderings (SOV-to-SVO);
many Japanese  sentences are difficult to be
 translated  into English monotonically.
Besides that, 
the test set has only one single reference and does not cover many possible expressions.
Results show that filtered Pseudo-Refs still improve the translation quality (Tab.~\ref{tab:bleu-jp}), and reduces anticipation (Fig.~\ref{fig:k-anticipation}) and hallucination (Tab.~\ref{tab:bleu-jp}).

\vspace{-0.2cm}
\section{Related Work}
\vspace{-0.2cm}

In the pre-neural statistical MT era, there exist several efforts using source-side reordering
as a preprocessing step for full-sentence translation \cite{collins+:2005,galley+:2008,xu+:2009}.
Unlike this work, they rewrite the source sentences.
But in the simultaneous translated scenario, 
the source input is incrementally revealed and unpredictable.
\citet{zheng2018multi} propose to improve full sentence
translation by generating pseudo-references
from multiple gold references, while our work does not
require the existence of {\em multiple} gold references and is
designed for simultaneous translation.

This work is closely related to the work of  \citet{he+:2015},
which
addresses the same problem but only in the special case of Ja$\goesto$En translation,
and uses handwritten language-specific syntactic transformations rules to rewrite the original reference into a more monotonic one.
By comparison, our work is much more general in the following aspects: 
(a) it is not restricted to any language pairs;
(b) it does not require language-specific grammar rules or syntactic processing tools;
and (c) it can generate pseudo-references with a specific policy according to the requirement of latency.

\vspace{-0.1cm}
\section{Conclusions}
\vspace{-0.2cm}

We have proposed a simple but effective method to generate more monotonic pseudo references for simultaneous translation.
These pseudo references cause fewer anticipations 
and can substantially improve simultaneous translation quality. 

\vspace{-0.2cm}
\section*{Acknowledgements}
\vspace{-0.2cm}
This work is supported in part by NSF IIS-1817231 and IIS-2009071 (L.H.).

\bibliographystyle{acl_natbib}
\interlinepenalty=10000 

\balance
\bibliography{nlp}


%
%
%
%
%
%

\clearpage

\appendix

\onecolumn

\section{Appendices}

\setcounter{figure}{0}
\renewcommand{\thefigure}{A\arabic{figure}} 

\setcounter{table}{0}
\renewcommand{\thetable}{A\arabic{table}}

  \begin{figure*}[!ht]\centering
    \resizebox{!}{1.1cm}{
    \setlength{\tabcolsep}{1.5pt}
    \renewcommand{\arraystretch}{1}
    \begin{tabu}{c | l l l l l l l l l l l l l l l l l l l}
    \toprule
    \rowfont{\small}
    & \textit{wǔjiǎodàlóu}  & \textit{méiyǒu} & \textit{xuānbù}  & \textit{xīn} & \textit{de} & \textbf{\textcolor{red}{\textit{fāshè}}} & \textbf{\textcolor{red}{\textit{rìqí}}} &。\\[-0.2cm]
    \tabincell{c}{Training\\Source Input} & 五角大楼 & 没有  &宣布 & 新 & 的  & \textbf{\textcolor{red}{发射}} & \textbf{\textcolor{red}{日期}} & 。 &\\[-0.2cm]
    & pentagon & not & announce & new & 's  & \textbf{\textcolor{red}{launch}}  & \textbf{\textcolor{red}{date}} &  &  &  & \\
  
    \midrule
    Gold Training-Ref & & & & no & new  & \textbf{\textcolor{red}{launch}} & \textbf{\textcolor{red}{date}} & was announced  by  the  pentagon . \\
    \midrule
    wait-3 Pseudo-Ref & & & & the & pentagon & has &not & announced  a  new  \textbf{\textcolor{red}{launch}}   \textbf{\textcolor{red}{date}} . \\
    \bottomrule
    \end{tabu}
    }
    \caption{The training reference uses passive voice while the source sentence uses active voice. 
    This kind of problem often appears in sentences with
    ``there be" (e.g. Fig.~\ref{fig:there}).
    The generated Pseudo-Ref can avoid anticipation by 
    keeping the active voice as the source sentence.}
    \label{fig:advance}
  \end{figure*}

  \begin{figure*}[!ht]\centering
    \resizebox{!}{1.1cm}{
    \setlength{\tabcolsep}{1.5pt}
    \renewcommand{\arraystretch}{1}
    \begin{tabu}{c | l l l l l l l l l l l l l l l l l l l}
    \toprule
    \rowfont{\small}
    & \textit{liǎng}  & \textit{guó} & \textit{jīngmào} & \textit{hézuò} & \textit{cúnzài} & \textit{zhe} & \textbf{\textcolor{red}{\textit{hěn}}} & \textbf{\textcolor{red}{\textit{dà}}} &\textbf{\textcolor{red}{\textit{de}}} &\textbf{\textcolor{red}{\textit{qiánlì}}} &。\\[-0.2cm]
    \tabincell{c}{Training\\Source Input} & 两 & 国  &经贸 & 合作 & 存在  & 着 & \textbf{\textcolor{red}{很}} & \textbf{\textcolor{red}{大}} & \textbf{\textcolor{red}{的}} & \textbf{\textcolor{red}{潜力}} &。\\[-0.2cm]
    &two  & country &economic trade & corperation & exist  &   & \textbf{\textcolor{red}{very}}  & \textbf{\textcolor{red}{big}} & \textbf{\textcolor{red}{'s}} & \textbf{\textcolor{red}{potential}}  &. \\
  
    \midrule
    Gold Training-Ref & & & & there & is  & \textbf{\textcolor{red}{very}} & \textbf{\textcolor{red}{great}} & \textbf{\textcolor{red}{potential}} & for & economic & and & trade  cooperation between  the  two  countries . \\
    \midrule
    wait-3 Pseudo-Ref & & & & the & economic  & and & trade & cooperation & between & the & two & countries  has  \textbf{\textcolor{red}{great}}  \textbf{\textcolor{red}{potential}} . \\
    \bottomrule
    \end{tabu}
    }
    \caption{A similar example in which the pseudo-reference
    avoids the anticipation brought by the
    ``there be'' phrase in the gold reference.}
    \label{fig:there}
  \end{figure*}

  \begin{figure*}[!ht]\centering
    \resizebox{!}{1.1cm}{
    \setlength{\tabcolsep}{1.5pt}
    \renewcommand{\arraystretch}{1}
    \begin{tabu}{c | l l l l l l l l l l l l l l l l l l l}
    \toprule
    \rowfont{\small}
    & \textit{dàn}  & \textit{xiéyì} & \textit{hái} &\textit{xūyào} & \textit{dédào} & \textit{sūdān} & \textit{nèigé} & \textit{de} &\textbf{\textcolor{red}{\textit{pīzhǔn}}} \\[-0.2cm]
    \tabincell{c}{Training\\Source Input} & 但 & 协议  &还 & 需要 & 得到  & 苏丹 & 内阁 & 的 & \textbf{\textcolor{red}{批准}} &。\\[-0.2cm]
    &but & agreement &also &need & get  & sudan  & cabinet & 's & \textbf{\textcolor{red}{approval}} & . & \\
  
    \midrule
    Gold Training-Ref & & & & but & the  & agreement & still & needs & \textbf{\textcolor{red}{approval}} & by & the & sud@@ &anese &cabinet &.\\
    \midrule
    wait-3 Pseudo-Ref & & & & but & the  & agreement & still & needs & to & be & \textbf{\textcolor{red}{approved}} & by & the & sud@@ &anese &cabinet &. \\
    \bottomrule
    \end{tabu}
    }
    \caption{
    The generated Pseudo-Ref avoids anticipation by adding a preposition ``to''. 
    }
    \label{fig:adding}
  \end{figure*}

  \begin{figure*}[!h]\centering
    \resizebox{1.0\linewidth}{!}{
    \setlength{\tabcolsep}{1.5pt}
    \renewcommand{\arraystretch}{1}
    \begin{tabu}{c | l l l l l l l l l l l l l l l l l l l l l l}
    \toprule
    \rowfont{\small}
    & \textit{wǒmen}  & \textit{de} & \textit{xīnwén} & \textit{méitǐ} & \textit{nénggòu} & \textit{dédào} & \textit{rénmín} & \textit{de} &\textit{xìnrèn} &, & \textbf{\textcolor{red}{\textit{gēnběn}}} & \textbf{\textcolor{red}{\textit{yuányīn}}} &\textit{jiù} & \textit{zài} &\textit{zhèlǐ} &.\\[-0.2cm]
    \tabincell{c}{Training\\Source Input} & 我们 & 的  &新闻 & 媒体 &能够  & 得到 & 人民 & 的 & 信任 &, &\textbf{\textcolor{red}{根本}} &\textbf{\textcolor{red}{原因}} &就 &在 &这里 &。\\[-0.2cm]
    &we  & 's &news &media & can  & get  & people & 's & trust &,  &\textbf{\textcolor{red}{fundamental}} & \textbf{\textcolor{red}{reason}} & that & on & this &.\\
  
    \midrule
    Gold Training-Ref & this & is  & the & \textbf{\textcolor{red}{fundamental}} & \textbf{\textcolor{red}{reason}}  & why & our &news & media &can & be & trust & by &the &people &.\\
    \midrule
    wait-3 Pseudo-Ref &&&& our & news & media & can & obtain  & the & trust & of & the & people &, & the & \textbf{\textcolor{red}{fundamental}}  \textbf{\textcolor{red}{reason}}  for  this . \\
    \midrule
    wait-5 Pseudo-Ref &&&&&& our & news & media & can & win & the & trust & of & the & people,  &and  this is the  \textbf{\textcolor{red}{fundamental}}  \textbf{\textcolor{red}{reason}} . \\
    \bottomrule
    \end{tabu}
    }
    \caption{
    Comparisons of Pseudo-Refs using different \waitk policies.
    These examples also show the trade-off between latency
    and fluency of pseudo-references.
    }
    \label{fig:diff}
  \end{figure*}

\end{CJK}
\end{document}